\documentclass[10pt,twocolumn,letterpaper]{article}

\usepackage[accsupp]{axessibility}
\usepackage{cvpr}              % To produce the CAMERA-READY version
\usepackage{graphicx}
\usepackage{amsmath}
\usepackage{amssymb}
\usepackage{booktabs}

\usepackage[pagebackref,breaklinks,colorlinks]{hyperref}

\usepackage[capitalize]{cleveref}
\crefname{section}{Sec.}{Secs.}
\Crefname{section}{Section}{Sections}
\Crefname{table}{Table}{Tables}
\crefname{table}{Tab.}{Tabs.}

\def\cvprPaperID{6} % *** Enter the CVPR Paper ID here
\def\confName{CVPR}
\def\confYear{2022}

\begin{document}

\title{GraphWalks: Efficient Shape Agnostic Geodesic Shortest Path Estimation}

\author{Rolandos Alexandros Potamias$^1$, Alexandros Neofytou$^2$, Kyriaki Margarita Bintsi$^1$, Stefanos Zafeiriou$^1$\\
$^1$Imperial College London, $^2$OCM Digital Media\\
{\tt\small \{{r.potamias,m.bintsi19,s.zafeiriou}\}@imperial.ac.uk}, {\tt\small alex@orangeclickmedia.com}  
% \and
% Alexandros Neofytou\\
% OCM Digital Media\\
% {\tt\small alex@orangeclickmedia.com} 
% \and
% Kyriaki-Margarita Bintsi\\
% Imperial College London \\
% {\tt\small m.bintsi19@imperial.ac.uk}  
% \and 
% Stefanos Zafeiriou \\
% Imperial College London \\
% {\tt\small s.zafeiriou@imperial.ac.uk} 
}
\maketitle

%%%%%%%%% ABSTRACT
\begin{abstract}
Geodesic paths and distances are among the most popular intrinsic properties of 3D surfaces. Traditionally, geodesic paths on discrete polygon surfaces were computed using shortest path algorithms, such as Dijkstra. However, such algorithms have two major limitations. They are non-differentiable which limits their direct usage in learnable pipelines and they are considerably time demanding. To address such limitations and alleviate the computational burden, we propose a learnable network to approximate geodesic paths. The proposed method is comprised by three major components: a graph neural network that encodes node positions in a high dimensional space, a path embedding that describes previously visited nodes and a point classifier that selects the next point in the path. The proposed method provides efficient approximations of the shortest paths and geodesic distances estimations. Given that all of the components of our method are fully differentiable, it can be directly plugged into any learnable pipeline as well as customized under any differentiable constraint. We extensively evaluate the proposed method with several qualitative and quantitative experiments. 
\end{abstract}

%%%%%%%%% BODY TEXT 
\section{Introduction}
\label{sec:intro}
Geodesic paths arise in several fields across 3D computer vision,  computer graphics and digital geometry with a broad amount of applications ranging from mesh morphing \cite{potamias2020learning} to robotics and circuit design \cite{673630}. A geodesic path can be considered any ``straight" curve that connects two points on a surface \cite{crane2020survey}. It is an intrinsic property of the 3D shape that remains unaffected by non-rigid deformations providing an expressive shape descriptor \cite{hopenfeld2007geodesic,shamai2017geodesic,xiang2021walk}. In a similar manner, geodesic distance can be defined as the distance of two points measured on the curved surface. Compared to Euclidean distance, geodesic distance is usually assumed to be a nonlinear function between the surface points. Traversal on different parts of a 3D object requires structural shape information that Euclidean distance usually fails to describe providing highly misleading results \cite{potamias2021revisiting,Potamias_2022_CVPR}. The computation of geodesic paths and distances requires careful consideration of the intrinsic surface properties to embody the influence of curvature. Although several techniques have attempted to embed intrinsic properties of the points and directly calculate geodesic distances on the continuous surface domain, in most of the 3D computer vision and graphics tasks, usually, such computations are made on discretized polyhedral surfaces, i.e. meshes. 

Given that 3D meshes can be considered as graphs embedded in the 3D Euclidean space, geodesic paths on discrete meshes are usually calculated using shortest path algorithms. Shortest path calculation between graph nodes lies in the core of many computer science algorithms and has been extensively studied with applications ranging from road networks and urban transportation \cite{lefebvre2007fast,wu2012shortest} to path planning \cite{das2018embodied}. Traditionally, algorithms for shortest paths calculation such as  \textbf{breadth-first-search (BFS)}$, \mathbf{A^*}$ \cite{hart1968formal}, \textbf{Dijkstra} and \textbf{Bellman-Ford} suffer from diminishing performance when scaling to significantly larger graphs. Due to the extreme amount of applications requiring detection of shortest path several extensions of the aforementioned algorithms have been developed to reduce the computational and storage requirements. Additionally, several approximation methods have been proposed \cite{Rattigan2006,Gubichev2010} to address the shortcomings of the traditional methods. Recently, several methods \cite{rizi2018shortest,qi2020learning} exploited the representation capabilities to estimate shortest path distances using deep learning. However, they rely on graph-specific node embeddings, such as Node2Vec \cite{grover2016node2vec} or DeepWalk \cite{perozzi2014deepwalk}, and can not be generalized to different graph topologies without re-training. Additionally, the aforementioned neural approximation methods only provide shortest path distances between graph nodes, thus the user can not directly infer the path connecting those nodes. In contrast, the shortest path estimation problem poses as a superordinate given that the shortest path distance can be easily computed using the weighted lengths of the predicted path. 

In this study, we propose the first, to the best of our knowledge, neural-based geodesic path estimator that is fully differentiable with respect to its inputs. The proposed methods is equipped with a graph neural network (GNN) with a novel graph message passing operator to encode the topology priors of the surface. The path is predicted using an autoregressive fashion, where the current path state is encoded using a momentum paradigm \cite{ioffe2015batch}. The proposed method outperforms all baseline methods in geodesic distance estimation and enables a path prediction speedup of up to 9 times. Finally, we showcase that the proposed method can accurately estimate geodesic paths and distances on unstructured point clouds.

\section{Related Work}
\subsection{Shortest Path}
\label{sec:shortestpath}
Finding the shortest path between a pair of a nodes with the purpose of minimizing the sum of edge weights is a problem that has been studied a lot in graph theory. Most of the existing shortest-path algorithms can be divided in two main categories, namely single-source shortest-path (SSSP), and all-pairs shortest-path (APSP). The algorithms may generate exact or approximate paths. While exact paths would be the optimal case, they are not always feasible to find, especially in large graphs. The APSP problem aims to extract the shortest paths between every possible pair of source $\mathcal{V}$ and destination $\mathcal{V'}$ nodes in the graph. The most popular APSP algorithm for both directed and undirected graphs is Floyd-Warshall \cite{10.1145/367766.368168} with a complexity of $\mathcal{O}(|\mathcal{V}|^3)$, where $|\mathcal{V}|$ is the number of the nodes. Another generalization of the shortest path problem is the SSSP problem, which consists of finding the shortest paths between a given vertex $\mathcal{V}$ and all other vertices in the graph. A standard choice for unweighted graphs is BFS \cite{cormen01introduction}, which, however, scales poorly and is proven impractical for today's massive graphs. Dijkstra’s algorithm \cite{DIJKSTRA1959} is used to solve the SSSP problem in both directed and undirected graphs with non-negative edge weights with a complexity of  $\mathcal{O}((|\mathcal{V}|+|\mathcal{E}|)\log{}|\mathcal{V}|)$ for the original algorithm, where $|\mathcal{V}|$ is the number of the nodes and $|\mathcal{E}|$ is the number of the edges, while the fastest to date version of the algorithm achieves a complexity of $\mathcal{O}(|\mathcal{E}|+|\mathcal{V}|\log{}|\mathcal{V}|)$ using a Fibonacci heap instead of the original min-priority queue \cite{10.1145/28869.28874,10.1145/50087.50096}. This is asymptotically the fastest known SSSP algorithm for the mentioned kind of graphs, with many implementations to use case-based optimizations which are out of the scope of this paper. An extension to the Dijkstra's algorithm is $\mathcal{A^*}$ \cite{hart1968formal}, a simple goal-directed algorithm that falls under the category of the best-first search algorithms. $\mathcal{A^*}$ finds the shortest path faster than Dijkstra, by using an admissible heuristic function. While the estimation of the exact shortest paths would be the optimal case, they are not always feasible to find, especially in very large graphs. In many cases, the estimation of an approximate path between two nodes is adequate. For this purpose, methods that include heuristic strategies \cite{Rattigan2006} and landmark strategies \cite{Gubichev2010} has been proposed. A Center Distance to Zone (CDZ) has been suggested in \cite{Tang2011} that considers 10\% of the nodes as central nodes and uses the Dijksta's algorithm in order to find the shortest path between central nodes. Although the CDZ algorithm manages to approximate the distance pretty accurately in some cases, it suffers from poor scaling with performance to drop on modern scale graphs. Several studies extended the notion of path approximation by modeling stochastic paths with edges that follow a probabilistic distributions instead of static costs \cite{chen2005path,lim2013practical,chen2021finding,mani2021shortest}.  However, in many practical scenarios there is no guarantee that the obtained random variables of edges are independent of each other \cite{yang2014stochastic}.  Many modifications of $\mathcal{A^*}$ have been implemented that combine the technique introduced in the $\mathcal{A^*}$ algorithm with the use of landmarks \cite{Goldberg2005ComputingPS,Gutman04}. A fast approximate of the real distance was suggested in \cite{10.1145/1645953.1646063}, that uses a subset of nodes as landmarks and estimates the distance of each node from each landmark. This however creates the problem of the correct choice of landmarks, which proved crucial for the accuracy of the estimation. More recently, machine learning techniques have been incorporated in the research of shortest path estimation. The authors in \cite{feijen2021using} propose an MLP to support and guide Dijkstra’s algorithm to reduce its search space. For an extensive survey of the filed we refer the reader to \cite{gallo1988shortest,foead2021systematic}. 

\subsection{Geodesic Distances}
Geodesic distance metrics prove a key ingredient to a vast amount of application ranging from representation learning \cite{xiang2021walk} to graph clustering \cite{pizzagalli2019trainable}. Calculation of geodesic distance can be achieved using two different approaches, based on the perspective of the 3D structure. Methods in the first approach consider 3D polygon meshes as a graph structure and calculate the geodesic distance as sum of the shortest path segments, using traditional shortest path algorithms as reviewed in Section \ref{sec:shortestpath}. In particular, an extension to Dijkstra algorithm, Fast Marching Method, along with several modifications of it \cite{kimmel1998computing,sethian1999level,bronstein2006efficient}, remains one of the most popular choices for geodesic distance computation on discretized surfaces.  On the contrary, methods on the second category calculate geodesic distances using algorithms from differentiable geometry under the assumption of a discretized differentiable surface. Although Euclidean distance seems a natural choice, the implicit assumption of linear geometric structure provides a misleading estimate for how far is a point in the nonlinear structure of 3D shapes \cite{bose2011survey}. In \cite{li2019geodesic}, the authors proposed a sphere piece fitting method to estimate local distances and calculate the geodesic distance through paths over spheres. Recently, Sharp \textit{et al.} \cite{sharp2020flipout} proposed a modification of Delaunay flip algorithm to estimate geodesic paths and distances by flipping edges of the mesh. However, such algorithms do not always produce actual shortest paths, hence cannot guarantee accurate geodesic distances. Apart from optimization based methods, there has been an attempt to model shortest path distances using neural networks. The authors of \cite{rizi2018shortest} proposed to utilize Node2Vec \cite{grover2016node2vec} embeddings to approximate shortest path distances using training node pairs obtained from pre-computed actual shortest path distances on several graph landmarks as in \cite{zhao2010orion}.  In a similar manner, Qi \textit{et al.} \cite{qi2020learning} trained a neural network to predict vertex embeddings along with a MLP to regress distance values. However, Node2Vec embeddings as well as the embedding layer proposed in \cite{qi2020learning}, can not be generalized to different graph topologies and thus they are impractical in 3D real-world application.  For additional resources on geodesic paths and distances we refer the reader to extensive surveys \cite{mitchell2000geometric,bose2011survey}.

% Approximation algorithms are compared according to their approximation ratio (or approximation factor) k. An algorithm that finds approximations to a geodesic path with approximation ratio k returns a path of length at most k times the exact geodesic path.

% Knowledge of the distances to the landmarks,
% together with the triangle inequality, typically allows one to
% compute approximate distance between any two nodes in O(l)
% time, where l is the number of landmarks. However, their results produce relatively high error rates, which
% limits the types of applications it can serve.

%  But these methods need high space cost. For some applications, such as finding nearest point of interest (POI) for travel recommendations, which only need the approximate distances.

\begin{figure*}[!ht]
    \centering
    \includegraphics[width=\linewidth]{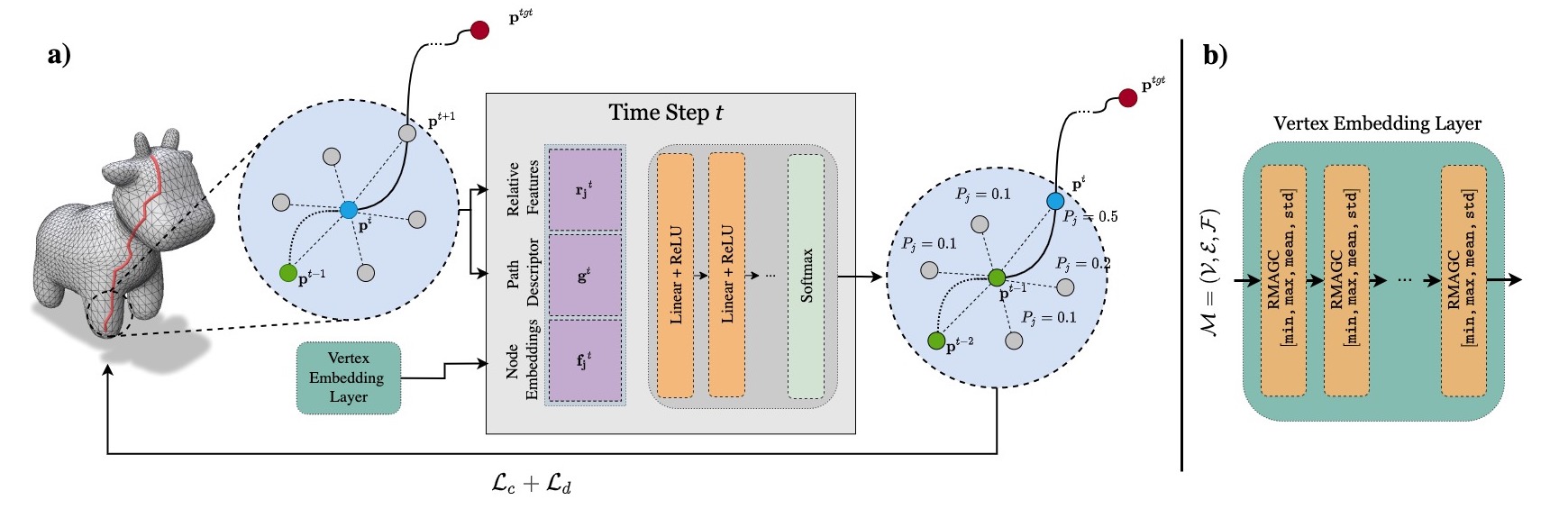}
    \caption{Architecture of the proposed method. a) At each time $t$, the Point Selector module decides the next point in the path sequence given the node embeddings (Sec. \ref{sec:vel}), the path descriptor (Sec. \ref{sec:descr}) and the neighborhood relative point features (Sec. \ref{sec:pointselector}). b) Node embeddings produced by the Vertex Embedding Layer (green) introduce a bias to the neighborhood features (purple) and assist the Point Selector decide the next point in the path sequence.}
    \label{fig:my_label}
\end{figure*}

\section{Method} 
Given a mesh $\mathcal{M}$=$(\mathcal{V},\mathcal{E},\mathcal{F})$ with vertices $\mathcal{V}$, edges $\mathcal{E}$ and faces $\mathcal{F}$, the proposed method aims to accurately estimate the geodesic path between two points $\mathbf{v}_{str}, \mathbf{v}_{tgt} \in \mathcal{V}$ on the surface $\mathcal{S}$. It takes as input the current point $\mathbf{p}^t$, its neighborhood $\mathcal{N}(\mathbf{p}^t)$ along with the current path descriptor $\mathbf{g}^{t}$ and predicts at every time step $t$ the next point in the path sequence using an autoregressive manner: 
\begin{equation}
    \mathbf{p}^{t+1}, \mathbf{g}^{t+1} = \mathcal{G} (\mathbf{p}^{t} , \mathcal{N}(\mathbf{p}^t), \mathbf{g}^{t})
\end{equation}
In the following section we describe the three main modules of the proposed method, namely: Vertex Embedding Layer, Path Descriptor and Point Selector along with the training objective utilized. 

\subsection{Vertex Embedding Layer}
\label{sec:vel}
Based on the observation that points close in space will share similar features \cite{ranjan2019asap}, we propose the use of graph neural network to encode the nodes of the 3D shape, that consumes relative position features rather than raw features as traditionally used in networks such as PointNet \cite{qi2017pointnet}. Using a GNN we exploit topology priors that are essential for the task of shortest path estimation. Specifically, assuming two points on two distinct fingertips that are considered close in the Euclidean space, these points are actually distant on the surface curve and consequently in the graph domain. Thus, the scope of Vertex Embedding Layer (VEL) is to assign topology aware features node, compared to topology agnostic networks such as PointNet \cite{qi2017pointnet}. We model VEL using a stack of graph convolution layers as defined bellow. 

\subsubsection{Relative Multi-Aggregation Graph Convolution}
Given that node embeddings are solely dependent on the expressive power of the GNN, we attempt to structure a graph operator that can effectively describe curved surfaces such as 3D meshes. The limitation of GNNs to expressively describe neighborhood information using a single aggregation function was initially explored in \cite{corso2020principal}. Based on such observations, we propose an expressive graph convolution operator, that is able to model neighborhood structure without being affected from outliers and local extremas. In particular, motivated by the works of \cite{hamilton2017inductive,ranjan2020asap,corso2020principal}, we introduce Relative Multi-Aggregation Graph Convolution (RMAGC) layer, with a node feature update rule defined as: 
\begin{equation}
    \mathbf{f}^{(l)}_{i} = \left[||\square_{k \in \mathcal{N}_i} \mathbf{W_n} \left(\mathbf{f}^{(l-1)}_{i} - \mathbf{f}^{(l-1)}_{j}\right)\right]
\end{equation}
where $\mathbf{f}^{(l)}_{i}$ is the feature of node $\mathbf{v}_i$ in layer $l$, $\mathbf{W_n}$ layer parameters,  $\mathcal{N}_i$ the neighbors of node $\mathbf{v}_i$, $\square$ the aggregation function and $||$ the concatenation operator. Note that by using Cartesian point position features the vertex embedding layer can only learn node descriptors aligned with the training set shapes. In contrast, the proposed relative feature formulation enables vertex embeddings decomposition from the shape topology. In the proposed setting, we utilized four stacked RMAGC layers with four aggregation functions: $\mathtt{[mean,min,max,std]}$. 

\subsection{Path Descriptor}
\label{sec:descr}
Apart from the general graph structure and shape, we need to proper encode the path direction to avoid predictions that are not aligned with the current predicted path. In particular, loops around nodes along with misleading direction vectors can derail the path from the target point.  To avoid that, we encode the prefix of the path up to the current node $\mathbf{p}^t = \mathbf{v}_i$ using the momentum formulation inspired from \cite{ioffe2015batch}: 
\begin{equation}
    \beta = \mathtt{sigmoid}([\mathbf{g}^{t-1} || \mathbf{f}_i^{t}] )
\end{equation}
\begin{equation}
   \mathbf{g}^{t} = \beta \mathbf{f}_i^{t} + (1-\beta)\mathbf{g}^{t-1}
\end{equation}
where $\mathbf{g}^{t}$ is the path descriptor in time step $t$ and $\mathbf{f}_i^{t}$ the encoding of node $\mathbf{v}_i$ in time step $t$. 
\subsection{Point Selector}
\label{sec:pointselector}
The point selector module acts as a classification model that selects the next point in the path from the neighborhood of the current point $\mathbf{p}^t$. Given the vertex embeddings $\mathbf{F}$, the path descriptor $\mathbf{g}^t$, we formulate an expressive way to represent the current point's neighborhood so that to aid the point selector module to decide the best fit in the path. In order to do so, we utilize a set of relative to the target point features. In particular, we define the relative features of point $\mathbf{v}_j$ in the neighborhood of $\mathbf{v}_i$ as: 
\begin{equation}
\begin{split}
    \mathbf{r}_j = [\mathbf{v}_{tgt} - \mathbf{v}_i || \mathtt{cosine}(\mathbf{v}_{tgt} - \mathbf{v}_j, \mathbf{v}_j - \mathbf{v}_i) || \\  
    \mathtt{cosine}(\mathbf{v}_{tgt} - \mathbf{v}_j, \mathbf{v}_{tgt} - \mathbf{v}_{str}) || d(\mathbf{v}_j, \mathbf{v}_{tgt}) ]
\end{split}
\end{equation}
where $\mathbf{v}_{str}, \mathbf{v}_{tgt}$ are the starting and the target points, $\mathtt{cosine}(\cdot)$ denotes the cosine similarity function and $d(\cdot)$ the Euclidean distance operator. Similar to the vertex embedding layer, we use relative features to enable generalization to different shapes and topologies. We additionally utilize cosine similarity features to capture the path directions and their alignment with the directional vector between starting and ending points.  

Finally, the probability that point $\mathbf{v}_j$, in the neighborhood of $\mathbf{v}_i$, will be the next point in the path sequence is be defined as: 
\begin{equation}
    \mathcal{P}(\mathbf{p}^{t+1} = \mathbf{v}_j | p^{t} , \mathcal{M} ) = \mathtt{softmax(MLP(}[r_j || f_j || g^t] ) ) 
\end{equation}
where $\mathcal{M}$ is the mesh that $\mathbf{v}_i$ lies and MLP denotes a multilayer perceptron. In this paper we used a 5 layer MLP accompanied with ReLU activations.  
\subsection{Loss Functions} 
To train the proposed method we utilize a combination of two objective functions, namely the point selector cross entropy loss and a probabilistic geodesic distance loss. 
\subsubsection{Point Selection Classification Loss}
The main scope of the point selector classification loss is to maximize the probability that, in every step of the path, the correct next point is selected by the Point Selector. In order to do so, we make use of a binary cross entropy loss $\mathcal{L}_c$ to maximize the raw point probabilities of the Point Selector module according to the ground truth points in the path. 
Additionally, to enforce training efficiency we follow the teacher forcing strategy, where we feed the model with the ground truth path points after each prediction. 

\subsubsection{Probabilistic Geodesic Distance Loss}
Given that there is no guarantee for a unique shortest path between two points, we utilize an additional loss function to ensure that, in every step, the geodesic distance is minimum. With this formulation, we reinforce the network to make predictions that minimize the distance between the current point and the target vertex. In mathematical parlance, we define the probabilistic geodesic loss as: 
\begin{equation}
    \mathcal{L}_d = \sum_{x,y,z}({\sum_{j \in \mathcal{N}_i} p_j }\mathbf{v}_{ij} - \mathbf{v}_t)^2
\end{equation}
where $\mathbf{v}_t$ is the ground truth vertex in step $t$, $\mathbf{v}_{ij}$ the $j$-th neighbor of current vertex $\mathbf{v}_i$ and $p_{j}$ its corresponding selection probability as assigned by the Point Selector module.

\subsection{Beam Search Inference}
Trying to further aid the model in abstaining from commiting "short-sighted" errors during inference, we observed that the selection of the next node in the shortest path be compared with the selection of the next token in Natural Language Generation (NLG). Thus, inspired by the de facto algorithm used for decoding in NLG problems, we adjust the beam search algorithm to operate with node probabilities. Beam search is a form of pruned breadth-first search where the breadth is limited to $k \in \mathcal{Z}+$ \cite{meister2020if}. It is a time-synchronous approximate search algorithm \cite{stahlberg2019nmt} which increases total prediction step complexity by a constant multiplicative factor equal to the chosen breadth $k$. Experimentally we choose a breadth of $k=3$ beams as the optimal accuracy-complexity tradeoff.

\section{Experiments}
In this section we evaluate the shortest paths generated by the proposed framework. We initially examine the generated path convergence and geodesic distances produced by our method. Additionally, we evaluate performance speedup of our module compared to traditional methods and showcase the contribution of each model module using an ablation study.   

\textbf{Dataset.} 
To train our model we utilized the TOSCA dataset \cite{bronstein2008numerical} that contains 80 unique shapes under different non-rigid transformations. Motivated by the diversity of the shapes and the enormous amount of rough regions they include we considered TOSCA dataset as a perfect fit for the task of geodesic path modeling. We randomly selected 60 shapes for training and 20 unique shapes for testing. To generate the dataset, we randomly selected 500 pairs of vertices for each shape and generated the ground truth paths using Bellman-Ford algorithm. 

\textbf{Baselines.}
For comparison, we selected several shortest path estimation baselines with different properties. Initially, we implemented two baseline methods that follow the same inference structure as the proposed method, without however having any learnable module. In particular, at each time step of the shortest path estimation a point from the neighborhood of the current point is selected as the next point in the path based on its distance to the target end point (\textit{Euclidean Distance}) or its cosine similarity directional vector (\textit{Cosine Similarity}). To be more precise, the first method selects the point $v_j$ on the neighborhood of $v_i$ that is closer, in terms of euclidean distance, to the target point $v_{tgt}$,  whereas the second method selects the point that maximizes the cosine similarity,  $\mathtt{cos}(v_j - v_i , v_{tgt} - v_j)$.  Additionally, we compare the proposed method with the geodesic path estimation algorithm that is based on edge flips proposed in \cite{shamai2017geodesic}, the probabilistic shortest path estimation algorithm as described in \cite{lim2013practical}, along with the recently proposed learnable geodesic distance estimation method \cite{qi2020learning}. We avoid comparing with \cite{rizi2018shortest}, since it requires training Node2Vec embeddings for every single mesh topology, which cannot be obtained in an end-to-end framework.

\subsection{Geodesic Paths and Distances}
To assess the performance of the proposed method, we measure the geodesic distance error (GDE) on the test set shapes of TOSCA dataset. Note that the topologies we used to evaluate the model are not present in the training set. We measure the GDE between the predicted and the ground truth paths as: 
\begin{equation}
    \mathtt{GDE} = \frac{|d_{pr} - d_{gt}|}{d_{gt}}
\end{equation}
where $d_{pr}, d_{gt}$ are the predicted and the ground truth geodesic distances respectively. 

\begin{figure*}[!ht]
    \centering
    \includegraphics[width=\linewidth]{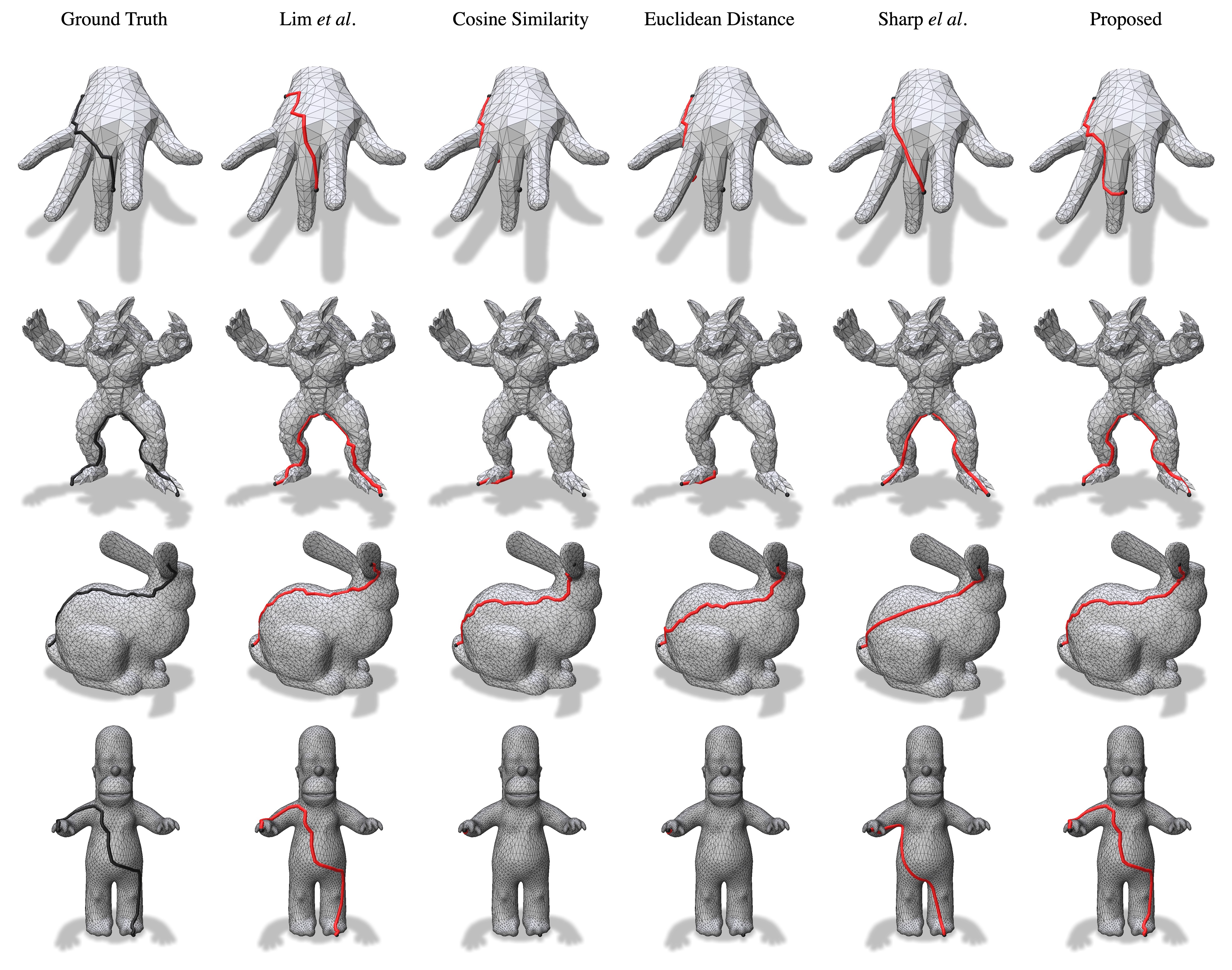}
    \caption{Qualitative comparison of the proposed and baseline methods. The ground truth shortest path is shown in black (left column) and the generated ones in red. Black dots denote start and end points.  }
    \label{fig:examples}
\end{figure*}

We also assess the convergence of each method in terms of steps required to reach the target point. In particular, for each method that predicts shortest path approximation, we measured the percentage of paths converged to the target point with up to 20\% more steps than the shortest path. 

As shown in Table \ref{tab:results}, the proposed method attains to achieve an increased convergence percentage along with a relatively small geodesic distance error, compared to the baseline methods. Significantly, apart from the method of Lim \textit{et al.} \cite{lim2013practical} all other baseline methods fail to consistently estimate paths that require similar steps to the ground truth shortest path. Additionally, as expected, methods that directly regress geodesic distances can not be easily adapted to unique mesh topologies and thus produce increased geodesic error. 
Such findings can be also qualitatively validated in Figure \ref{fig:examples}, where the proposed method is contrasted with the baselines under different shape topologies. It can be easily observed that the proposed method achieves to estimate accurate shortest path even in challenging shapes. In particular, in the case of the armadillo model (second row) as well as the Simpsons Homer model (fourth row), naive estimators such as the Cosine Similarity and the Euclidean distance get trapped in local minima and fail to converge. In contrast, the proposed method attains minimum geodesic distance error by selecting paths driven by the VEL and the path descriptor. Note the the method of Sharp \textit{et al.} \cite{sharp2020flipout}, breaks the mesh connectivity and thus produces paths that are not aligned with the original mesh topology. 

\begin{table}[]
    \caption{Quantitative evaluation of the proposed and baseline methods. }
    \label{tab:results}
    \centering
    \begin{tabular}{ l|c|c } 
    Method & Convergence (\%) & GDE \\ \hline
    Cosine Similarity & 38 & 0.39 \\ \hline
    Euclidean Distance & 42 & 0.35 \\ \hline
    Sharp \textit{et al.} \cite{sharp2020flipout} & 24 & 0.37 \\ \hline
    Lim \textit{et al.} \cite{lim2013practical} & 79 &  0.38 \\ \hline
    Qi \textit{et al.} \cite{qi2020learning}& - & 0.88  \\ 
      \hline
    \textbf{Proposed} & 
    \textbf{82} & \textbf{0.17} \\ 
     \hline \hline
    \end{tabular}
\end{table}

\begin{figure*}[!ht]
    \centering
    \includegraphics[width=\linewidth]{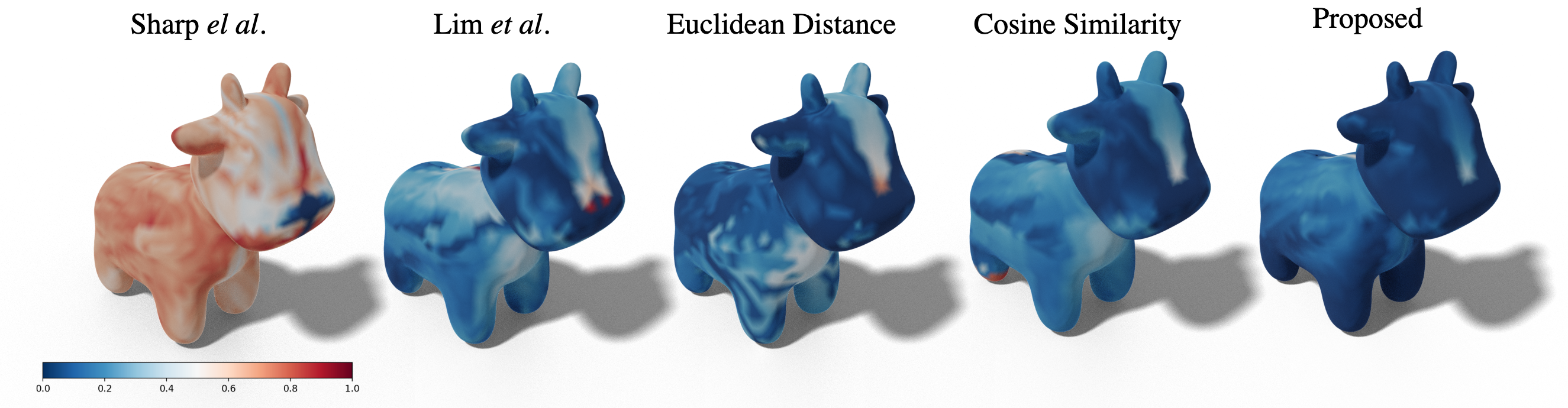}
    \caption{Geodesic distance error color-coded on the mesh surface. Blue color denotes small GDE whereas red color denotes increased error.  }
    \label{fig:colorcoded}
\end{figure*}

Additionally, we estimated the shortest paths from a single point to all points of the mesh and measured their geodesic error compared to the ground truth paths. In Figure \ref{fig:colorcoded}, we qualitatively assess the geodesic error of the generated geodesic paths. As can be easily seen, the proposed method generates paths that respect the intrinsic property of the mesh and achieves the smallest geodesic error across the whole mesh surface.  

\subsection{Runtime}
\begin{figure}[!h]
    \centering
    \includegraphics[width=\linewidth]{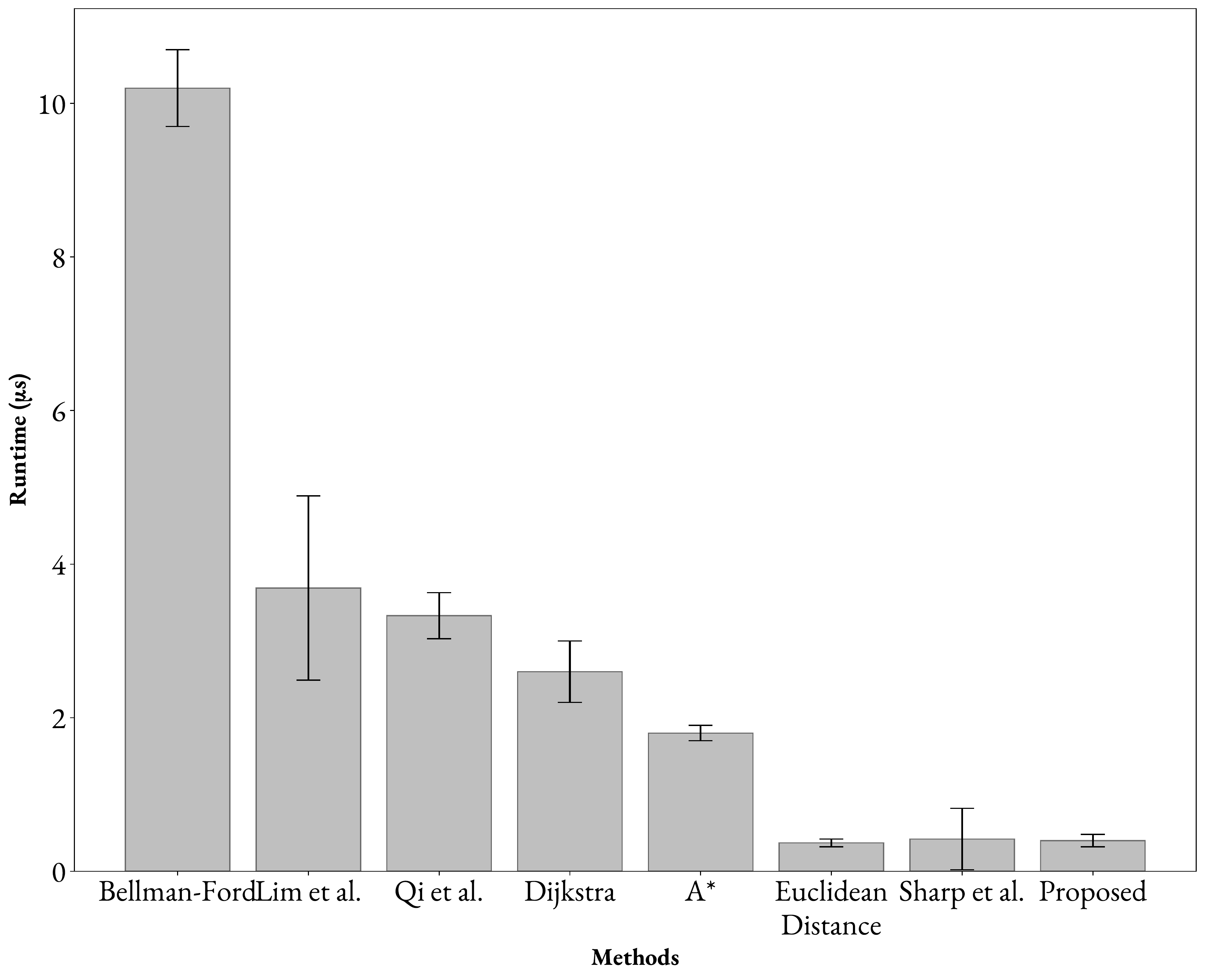}
    \caption{Runtime comparison between the proposed and baseline methods.}
    \label{fig:runtime}
\end{figure}

A main challenge of modern applications of shortest path and geodesic distance calculations is real-time execution of the algorithm. Apart from accurate and plausible solutions, a key property of a geodesic path estimation method is to diminish the runtime of traditional greedy algorithms. Taking advantage of the progress of modern GPUs, neural networks have become vastly popular for their precision as real-time approximators. To compare the proposed and the baseline methods we constructed an experiment with a total of 50 paths and measured the total runtime each method required to generate the solutions. Note that we have not segregated CPU from GPU methods since we are interested solely in the total runtime required. In Figure \ref{fig:runtime} we report the average runtimes on 20 executions along with their respective standard deviations. 
Regarding the implementation of choice for the traditional shortest path algorithms we selected for our baselines i.e. Bellman-Ford and Dijkstra, past research has aimed to optimize the algorithms' runtime on modern GPUs, as per \cite{ortega2013new,chaibou2015improving,Nayak2018ACS}. However, such optimized prototypes have not found their spot in modern data science workflows due to the difficulty of incorporating them into the codebases of modern data science libraries, a fact often attributed to the absence of code for reproduction of their implementation and experiments. Therefore, we use the immensely popular library SciPy \cite{virtanen2020scipy} as the source for our primitive algorithm baselines. While there is no explicit support of SciPy for GPU computing, its functions are highly optimized to run on modern CPUs as it uses Python wrappers around algorithms coded in C or C++.
As seen in Figure \ref{fig:runtime}, the proposed method runs on par with or faster than all compared baselines, both differentiable and purely algorithmic. Compared with traditional shortest path algorithms, our method scales down the runtime by up to 9 times, setting itself ready performance-wise for integration in broader learning frameworks. Furthermore, the proposed method attains to achieve efficient performance both in terms of GDE and convergence percentage as well as runtime execution compared to Euclidean Distance and Sharp \textit{et al.} \cite{sharp2020flipout} methods.

\subsection{Shortest Path Estimation on Point Clouds}
As an attempt to evaluate the performance of the proposed method in extreme cases where the explicit 3D shape topology is absent, we attempted to estimate shortest paths on unstructured point clouds. In particular, we fed to the model point clouds where the connectivity is attained using k-nearest neighbors (k-NN) and we compared the predicted paths with the ground truth paths, obtained from the original mesh connectivity. We structure the neighborhood of each point by using five nearest neighbors based on their Euclidean distance. 
\begin{figure}[!ht]
    \centering
    \includegraphics[width=\linewidth]{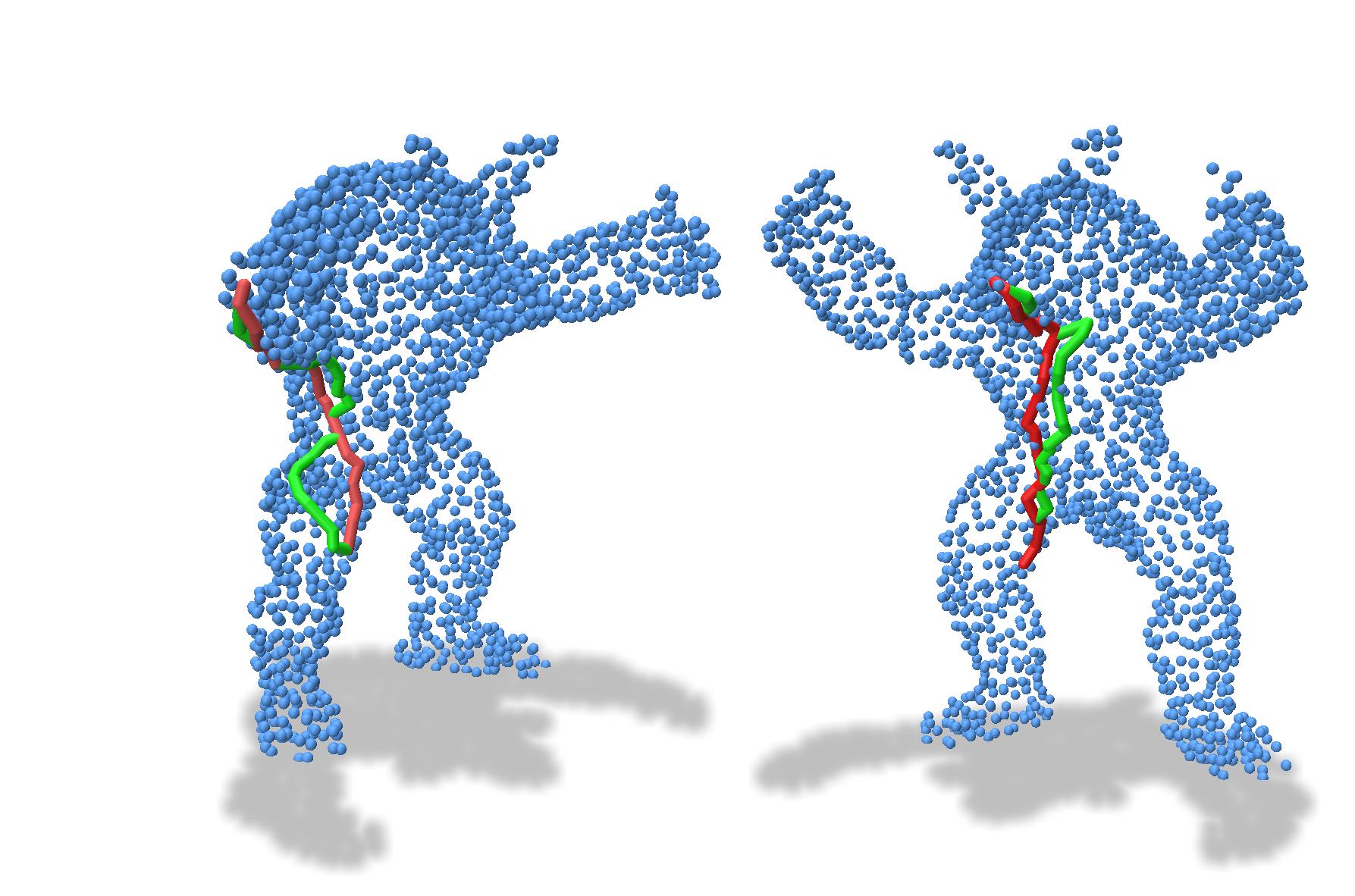}
    \caption{Generalization of the proposed method on an unstructured point cloud. Predicted geodesic paths (green) achieve a GDE of 0.21 on the armadillo 3D shape compared to the ground truth paths (red).}
    \label{fig:pointcloud}
\end{figure}
As can be seen in Figure \ref{fig:pointcloud}, the proposed method converges to the target point with a small amount of extra steps and achieves remarkable GDE compared to the ground truth paths.

\begin{figure}[!ht]
    \centering
    \includegraphics[width=\linewidth]{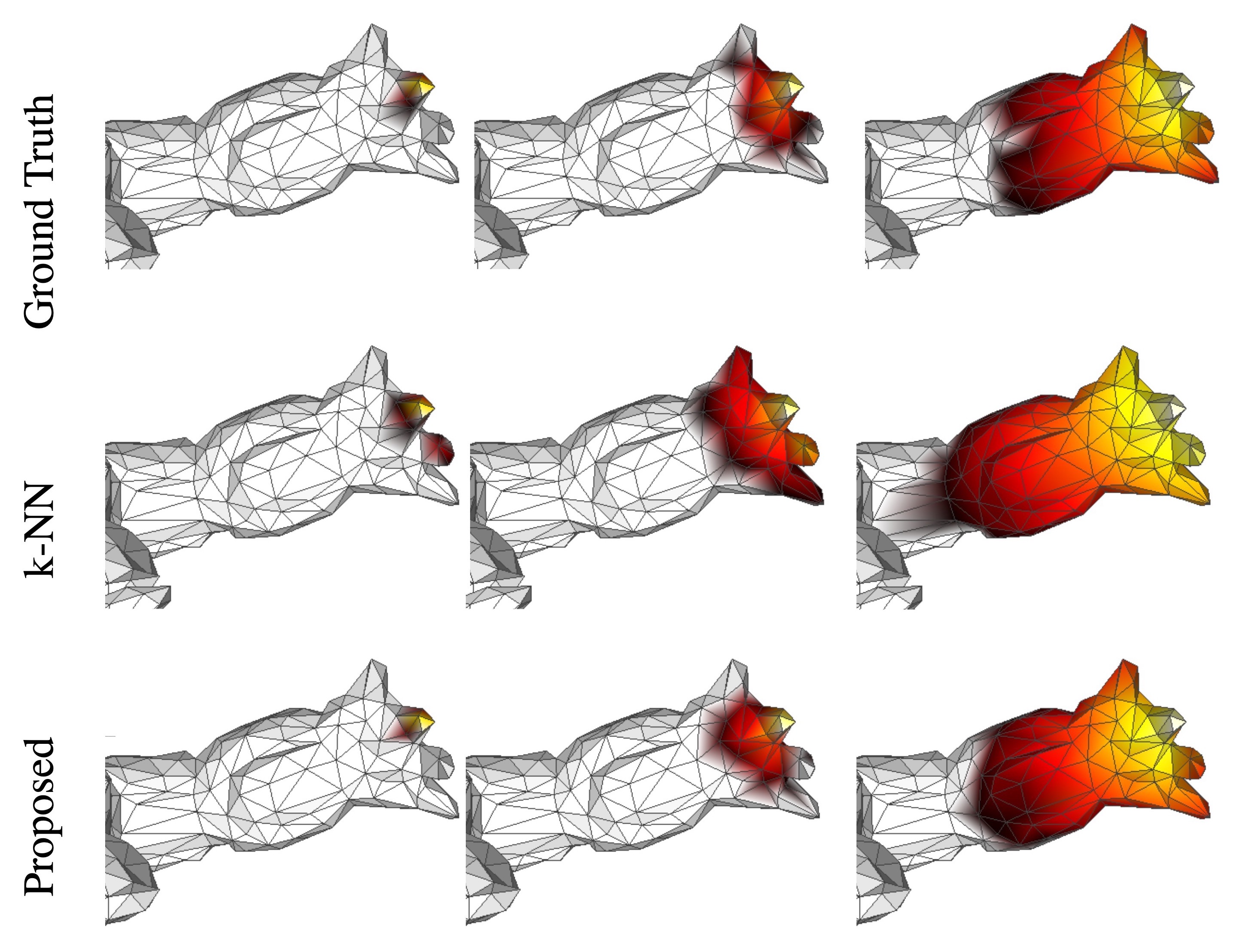}
    \caption{Color-coded neighborhood points estimation. From left to right we visualize points that are 0.1 , 0.2 and 0.5 away from the finger tip. Results are illustrated on top of the structured mesh for better visualization. }
    \label{fig:neighborhood_estimation}
\end{figure}

Additionally, we examined the case of point neighborhood estimation on the armadillo point cloud. In particular, using the k-NN connectivity as described earlier, we estimated the geodesic distances from a fingertip to all other points of the point cloud and retained the ones that are less than a given threshold. In Figure \ref{fig:neighborhood_estimation}, we compare the geodesic neighborhood of a finger tip point estimated by the proposed method and the k-NN method, that is the common practice in point cloud analysis task. It can be easily observed that using the proposed method we achieved geodesic neighborhood estimates that are very similar to the ground truth neighborhood of the point. On the contrary, k-NN method that is based purely on Euclidean point distances, fails to distinguish different fingers, which is particularly useful to tasks such as part segmentation. 

\subsection{Ablation Study}
We performed an ablation study to illustrate the effectiveness of the key components of the proposed method. In particular, we assessed the importance of both three major modules of the proposed network along with the impact of each loss functions to the method's performance. 
Experimental results on the test set are summarized in Table \ref{tab:ablation}. The importance of embedding node features and graph topology to estimate paths on the surface can be quantified with the removal of vertex embedding layer (VEL). As expected, the absence of VEL has a large impact on the path convergence leading to a significant decrease in performance. Similarly, a performance drop can be observed when we substitute relative features, as introduced in \ref{sec:pointselector}, with $(x,y,z)-$coordinate features. Using directly $(x,y,z)-$coordinate features the model can not generalize well to unseen figures and paths, whereas utilizing relative feature we model a, much easier, space of deformations, that aids network generalization capabilities. Experimental results also reveal that path descriptor is also important in the process of path estimation. Specifically, the momentum strategy introduced in the path descriptor boost the point selector to select points of the neighborhood that follow the pattern of the prefix path avoiding loops and derailed paths. 

Finally, the contribution of each loss function was examined by training the proposed network using each loss individually. The two losses can be considered as a direct and indirect way of training the network, respectively. Classification loss impacts the point selector generated probabilities directly based on the ground truth points whereas the geodesic loss indirectly guides network predictions to minimize the distance to the target point. Using only the geodesic loss without the classification loss, the model tends to imitate the Euclidean Distance baseline where decisions are made purely by the distance to the end goal. However, its contribution is not negligible given that it enforces training of certain parts of the network such as the path descriptor.

\begin{table}[]
    \caption{Ablation Study: Quantitative evaluation of the proposed method key components.}
    \label{tab:ablation}
    \centering
    \begin{tabular}{ l|c|c } 
    Method & Convergence (\%) & GDE \\ \hline
    \quad - w/o VEL & 57 & 0.35 \\ \hline
    \quad - w/o Path Descriptor   & 71 & 0.23 \\ \hline
    \quad - w/o Relative Features & 66 & 0.29  \\ \hline
    \quad - w/o Classification Loss & 53 &  0.34 \\ \hline
    \quad - w/o Geodesic  Loss & 78 & 0.24  \\ 
      \hline
    \textbf{Proposed - Full} & 
    \textbf{82} & \textbf{0.17} \\ 
     \hline \hline
    \end{tabular}
\end{table}

\section{Limitations and Conclusion}
In this paper, we presented the first, to the best of our knowledge, geodesic path estimator that is based on neural networks. Motivated by the lack of differentiable and real-time estimation methods along with the success of neural networks as approximators, we implemented a framework that learns to approximate the shortest path problem. In particular, we model the shortest path problem using an autoregressive model that given the current point, its neighboring points and the prefix of the path, decides the next points in the path sequence. The proposed method is comprised by three main modules namely the vertex embedding layer that generates node embeddings given the mesh topology, a path descriptor that models the prefix path encodings and the point selector that given such information chooses the next point in the geodesic path. We evaluate the proposed method under a test set of diverse shapes and we showcase its superiority compared to other baseline approximation methods. Additionally, we explore the generalization capabilities of the proposed method on unstructured point clouds and illustrate that it can be directly used to point clouds without being significantly affected by the lack of point connectivity. 

However, the proposed method is strongly attached to the approximation quality of the neural network, which sometimes limits the generated shortest paths. Specifically, in extreme scenarios that have not been explored in the training set and in cases where the shortest path is not obvious, the approximation quality of neural networks might be strictly bounded. To tackle such limitations, we plan to extend the proposed method to generate more robust and expressive features that will break the approximation burdens of neural networks. Finally, we believe that the rationale underlying the proposed method and the presented findings of this paper will not only benefit the 3D computer vision but also the geometric deep learning community.

%%%%%%%%% REFERENCES
{\small
\bibliographystyle{ieee_fullname}
\bibliography{egbib}
}

\end{document}